\documentclass{article}

\PassOptionsToPackage{colorlinks=true, linkcolor=blue, citecolor=blue, urlcolor=blue}{hyperref}

\usepackage[preprint]{neurips_2026}

\makeatletter
\def\@noticestring{}
\makeatother

\usepackage[utf8]{inputenc}
\usepackage[T1]{fontenc}
\usepackage{url}
\usepackage{booktabs}
\usepackage{amsmath}
\usepackage{amsfonts}
\usepackage{nicefrac}
\usepackage{microtype}
\usepackage{xcolor}
\usepackage{graphicx}
\usepackage{multirow}
\usepackage{subcaption}
\usepackage{textcomp}
\usepackage{hyperref}

\hypersetup{
    colorlinks=true,
    linkcolor=blue,
    citecolor=blue,
    urlcolor=blue
}

\title{Vera: Identity-Faithful Human Subject-to-Video Generation}

\newcommand{\appentry}[2]{%
    \noindent\hyperref[#2]{\textcolor{blue!70!black}{\textbf{#1}}}%
    \dotfill%
    \pageref{#2}\par
}

\newcommand{\appsubentry}[2]{%
    \noindent\hspace{1.8em}\hyperref[#2]{\textcolor{blue!70!black}{#1}}%
    \dotfill%
    \pageref{#2}\par
}

\renewcommand{\thefootnote}{\fnsymbol{footnote}}
\newcommand{\affmark}[1]{\raisebox{0.75ex}{\normalfont\scriptsize #1}}
\newcommand{\equalmark}{\raisebox{0.75ex}{\normalfont\scriptsize *}}

\author{%
\normalfont\mdseries
Yulong Xu\affmark{1}\thanks{Equal contribution.},
Xinyue Liu\affmark{1,2}\equalmark,
Shujuan Li\affmark{1,3},
Huafeng Shi\affmark{1},
Yan Zhou\affmark{1},\\
Jiwen Liu\affmark{1},
Xintao Wang\affmark{1},
Pengfei Wan\affmark{1},
Yu-Shen Liu\affmark{1},
Huaibo Huang\affmark{2}\thanks{Corresponding author.}\\
\affmark{1}Kuaishou Technology \quad
\affmark{2}Institute of Automation, Chinese Academy of Sciences \quad
\affmark{3}Tsinghua University\\
\texttt{xuyulong03@kuaishou.com} \quad
\texttt{liuxinyue2025@ia.ac.cn}
}

\begin{document}

\maketitle

\setcounter{footnote}{0}
\renewcommand{\thefootnote}{\arabic{footnote}}

\begin{abstract}
\label{sec:abs}

Subject-to-video (S2V) generation has made substantial progress in preserving reference subjects across diverse categories, yet generic subject consistency remains insufficient for human-centric generation. A video may appear globally consistent while identity-critical human details still drift across frames, poses, and interactions. This issue becomes more severe in multi-person scenarios, where incorrect identity-role binding leads to subject confusion, attribute swapping, and excessive copying of reference-specific appearance cues.
We propose Vera, a unified human-centric S2V framework for single- and multi-person generation. We first construct a million-pair identity-aligned human image-video dataset through person-level cross-clip retrieval, providing explicit identity correspondence and diverse references. Built on this dataset, Vera introduces two complementary designs. Identity-Focal Masked Supervision (IFMS) strengthens identity-aware learning with spatially focused supervision while reducing interference from irrelevant artifacts. Reference-Aware Layer-wise Attention (RALA) regulates how video tokens interact with reference identity cues in the DiT backbone, preserving stable identity anchors and enhancing layer-aware identity readout. Extensive experiments demonstrate that Vera improves human identity consistency, multi-person subject binding, and motion naturalness, while reducing identity confusion and excessive reference-image copying.
\end{abstract}
\section{Introduction}
\label{sec:intro}

In human-centered video generation, facial identity serves as the anchor of visual continuity. A person may move, turn, interact with others, or appear under different viewpoints, yet the viewer should still recognize the same individual throughout the video. This requirement is fundamental for personalized storytelling, digital human creation, advertising, and character-driven content generation. However, reproducing such identity permanence in generative video models remains challenging. Although recent diffusion-based and Diffusion Transformer video models have achieved remarkable photorealism and temporal coherence \cite{peebles2023scalable, yang2024cogvideox, polyak2024movie}, they often struggle to preserve human identity under dynamic conditions. As head pose, facial angle, occlusion, and scene context change, generated faces may gradually drift from the reference identity; in multi-person scenes, different identities may even become entangled.

Subject-to-video(S2V) generation provides a natural formulation for human-centered controllable video synthesis by conditioning the model on reference images and text prompts \cite{jiang2024videobooth, liu2025phantom, jiang2025vace, yuan2025opens2v}. Existing S2V and video personalization methods have demonstrated promising results in preserving general subjects, including humans, animals, objects, garments, and scenes \cite{chen2025multi, liang2025movie, huang2025conceptmaster}. However, human-centric S2V requires more than generic subject preservation. Maintaining a person is not merely preserving coarse appearance cues such as clothing, hairstyle, or body shape; it requires per-person facial identity to remain stable across frames, poses, interactions, and viewpoints. This distinction becomes especially important in a unified setting covering both single- and multi-person generation, where the model must correctly bind each reference identity to its corresponding textual role and generated trajectory.

Current approaches suffer from two primary limitations. First, they often optimize subject-level appearance consistency rather than fine-grained facial identity consistency. A generated video may appear globally consistent with the reference subject, while the small facial regions that determine identity still degrade or change over time. Identity-preserving video methods improve facial consistency by introducing dedicated identity encoders or control mechanisms \cite{he2024id, yuan2025identity, xue2025stand}, but they are typically designed under a single-identity assumption, where one reference identity corresponds to one generated subject. In contrast, multi-person S2V requires not only identity preservation, but also identity-role-trajectory binding: multiple reference identities must be correctly associated with multiple textual roles and generated trajectories. Stronger reference conditioning does not necessarily solve this issue. Naively injecting reference images into the video backbone can cause the model to treat reference views as static templates, leading to pose locking, view collapse, and over-preservation of reference-specific pose and appearance cues \cite{liu2025phantom, guo2026wildactor}. In multi-person scenes, such naive fusion further introduces identity-role misbinding, subject confusion, and attribute swapping \cite{huang2025conceptmaster}.

Second, there is a lack of large-scale training data with explicit identity correspondence for human-centric S2V. Recent datasets and benchmarks have advanced S2V generation at scale \cite{yuan2026opens2v}, and human-centric datasets have explored identity preservation under diverse viewpoints and motions \cite{guo2026wildactor}. However, generic image-video pairs are insufficient for learning stable facial identity, while in-pair references sampled from the same clip can encourage shortcut learning and copy-paste. For multi-person videos, the data construction problem becomes even harder: the dataset must distinguish different identities within a clip, match each identity to appropriate reference images, and preserve diversity across facial poses and appearances. Without such identity-aware supervision, models struggle to learn robust person-level correspondence under natural motions, occlusions, and interactions.

To address the above challenges, we propose Vera, a unified human-centric S2V framework for identity-faithful single- and multi-person generation. Vera is built upon a million-pair identity-aligned human image-video dataset constructed through person-level cross-clip retrieval. Instead of using reference frames from the target clip, we retrieve identity-matched references from other clips, preserving facial identity while introducing variations in pose, expression, illumination, background, and motion. This cross-clip design encourages identity correspondence learning beyond direct frame reconstruction, reducing shortcut learning, background leakage, and reference-induced over-copying of frame-specific reference cues. It also provides explicit person-level supervision for facial identity preservation and multi-person subject binding.

Leveraging this dataset, Vera introduces two simple yet effective designs tailored for identity preservation. First, Identity-Focal Masked Supervision (IFMS) reweights the training objective with region-aware masks, emphasizing identity-critical facial regions while suppressing irrelevant subtitle artifacts. This strengthens fine-grained identity supervision without changing the inference pipeline. Second, Reference-Aware Layer-wise Attention (RALA) regulates reference-video interaction in the DiT backbone. It preserves reference tokens as stable identity anchors in early layers and selectively enhances video-to-reference identity readout at identity-sensitive layers, improving fine-grained identity preservation while maintaining flexible cross-modal fusion.


In summary, our contributions are:
\begin{itemize}

\item We construct a million-pair identity-aligned human image-video dataset through person-level cross-clip retrieval, providing explicit identity correspondence for facial identity preservation and multi-person subject binding.
\item We propose Vera, incorporating Identity-Focal Masked Supervision (IFMS) and Reference-Aware Layer-wise Attention (RALA) to strengthen identity-aware learning and stabilize reference-video interaction.
\item Extensive comparisons with representative S2V baselines demonstrate the effectiveness of Vera in preserving facial identity, maintaining subject-role correspondence in multi-person scenes, and reducing over-reliance on reference images.
\end{itemize}

\section{Related Work}
\label{sec:related work}

\subsection{Video Generation Models}

Diffusion models have driven rapid progress in high-fidelity video generation. Early methods extend image diffusion models with temporal modeling, such as Make-A-Video~\cite{singer2022make}, Stable Video Diffusion~\cite{blattmann2023stable}, and AnimateDiff~\cite{guo2023animatediff}. More recently, scalable Transformer-based architectures, including DiT~\cite{peebles2023scalable} and rectified-flow formulations, have enabled large video foundation models such as CogVideoX~\cite{yang2024cogvideox}, HunyuanVideo~\cite{kong2024hunyuanvideo}, Wan~\cite{wan2025wan}, Open-Sora~\cite{zheng2024open}, and Movie Gen~\cite{polyak2024movie}. These models substantially improve visual quality, text-video alignment, and motion coherence, and have become strong backbones for controllable video synthesis.

However, existing video foundation models are mainly optimized for general-purpose generation rather than identity-sensitive human generation. While they can synthesize realistic scenes and coherent motions, they still struggle to preserve a specific person under pose changes, facial expressions, occlusions, and multi-person interactions. Vera addresses this gap by introducing human-centric supervision and reference conditioning tailored to facial identity preservation.

\subsection{Subject-Consistent Video Generation}
Subject-consistent generation synthesizes content that follows text prompts while preserving reference subjects. Image personalization methods, from optimization-based approaches such as Textual Inversion~\cite{gal2022image} and DreamBooth~\cite{ruiz2023dreambooth} to adapter-based methods such as IP-Adapter~\cite{ye2023ip}, InstantID~\cite{wang2024instantid}, and PuLID~\cite{guo2024pulid}, have shown strong reference-based controllability. Directly extending them to video, however, is insufficient due to the additional requirements of temporal consistency, motion naturalness, and viewpoint-robust identity preservation. Recent S2V methods, including VideoBooth~\cite{jiang2024videobooth}, DreamVideo~\cite{wei2024dreamvideo}, Tune-A-Video~\cite{wu2023tune}, PersonalVideo~\cite{li2025personalvideo}, Phantom~\cite{liu2025phantom}, VACE~\cite{jiang2025vace}, ConceptMaster~\cite{huang2025conceptmaster}, MAGREF~\cite{deng2025magref}, SkyReels-A2~\cite{fei2025skyreels}, MovieWeaver~\cite{liang2025movie}, and BindWeave~\cite{li2025bindweave}, improve subject preservation in single- and multi-subject scenarios. Nevertheless, they primarily target generic subject consistency, where coarse appearance can remain stable while facial identity still drifts.
Human-centric methods such as ID-Animator~\cite{he2024id}, ConsisID~\cite{yuan2025identity} focus on facial identity preservation, but are mostly limited to single-person settings or do not explicitly model identity-role binding in multi-person scenes. Vera complements these works by using identity-aligned cross-clip data, face-focused supervision, and reference-preserving attention to improve facial identity consistency, subject binding, and natural motion for unified single- and multi-person S2V.
\begin{figure*}[t]
  \centering
  \includegraphics[width=\linewidth]{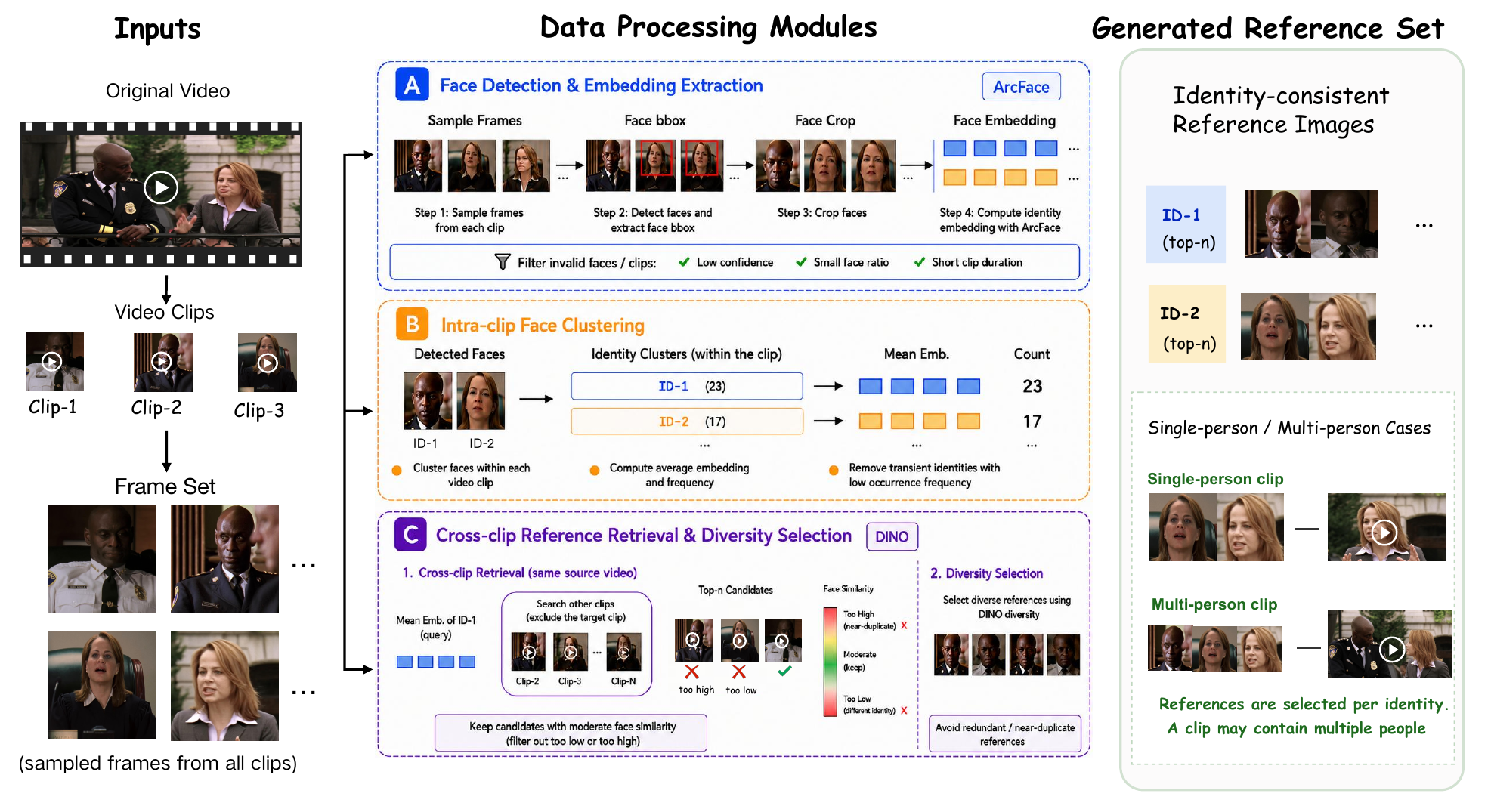}  
  \caption{ Data processing pipeline for identity-consistent video generation. The process includes video clipping, frame sampling, face detection and embedding extraction, intra-clip identity clustering, and diverse reference selection to build identity-aligned reference sets for single- and multi-person human S2V generation.}
  \label{fig:data}
\end{figure*}
\section{Methodology}
\label{sec:method}
In this section, we introduce the proposed \textbf{Vera} framework. We first present the identity-aligned data pipeline in Section~\ref{sec:1}. Next, we describe \textbf{Identity-Focal Masked Supervision (IFMS)} in Section~\ref{sec:2} and \textbf{Reference-Aware Layer-wise Attention (RALA)} in Section~\ref{sec:3}. Together, these designs enable robust facial identity preservation and subject binding for S2V generation. Fig.~\ref{fig:pipeline} provides a schematic overview of the Vera architecture.
\subsection{Data Pipeline}
\label{sec:1}
To support human-centric S2V generation, we construct a large-scale identity-aligned image-video dataset, as shown in Fig.~\ref{fig:data}. Each training sample consists of a target video clip and one or multiple reference images, where each reference image is explicitly associated with a specific person appearing in the video. This data format provides direct supervision for both facial identity preservation and multi-person subject binding.

The key principle of our data construction is cross-clip identity pairing. Instead of using reference frames sampled from the same target clip, which may lead to trivial reconstruction by copying pose, background, and layout cues, we retrieve identity-matched reference images from different clips within the same video source. This preserves the underlying facial identity while introducing natural variations in pose, expression, illumination, background, and motion, encouraging the model to learn identity consistency beyond frame-level appearance matching.

Specifically, we first segment raw videos into temporally coherent clips and extract face-level identity information from sampled frames. Within each clip, detected faces are grouped into person-level identity clusters, allowing us to identify the main subjects and remove unreliable or transient faces. We then perform cross-clip retrieval within the same video source by using the averaged face embedding of each identity cluster as the query. To ensure reliable yet non-trivial pairings, we retain retrieved candidates whose face similarity falls within a moderate range, filtering out low-similarity matches that may correspond to different identities and overly high-similarity matches that may indicate near-duplicate frames. Finally, for each valid identity, we select diverse reference images from the retrieved candidates to avoid redundant samples. In multi-person clips, this process is performed independently for each identity, resulting in person-specific reference sets aligned with different subjects in the target video.

Through this pipeline, we obtain around one million identity-aligned human image-video pairs. Compared with generic subject-pair data, our dataset provides explicit person-level correspondence and diverse cross-clip references, enabling scalable training for facial identity preservation, identity-role binding, and robust S2V generation.

\subsection{Identity-Focal Masked Supervision}
\label{sec:2}
\begin{figure*}[t]
  \centering
  \includegraphics[width=\linewidth]{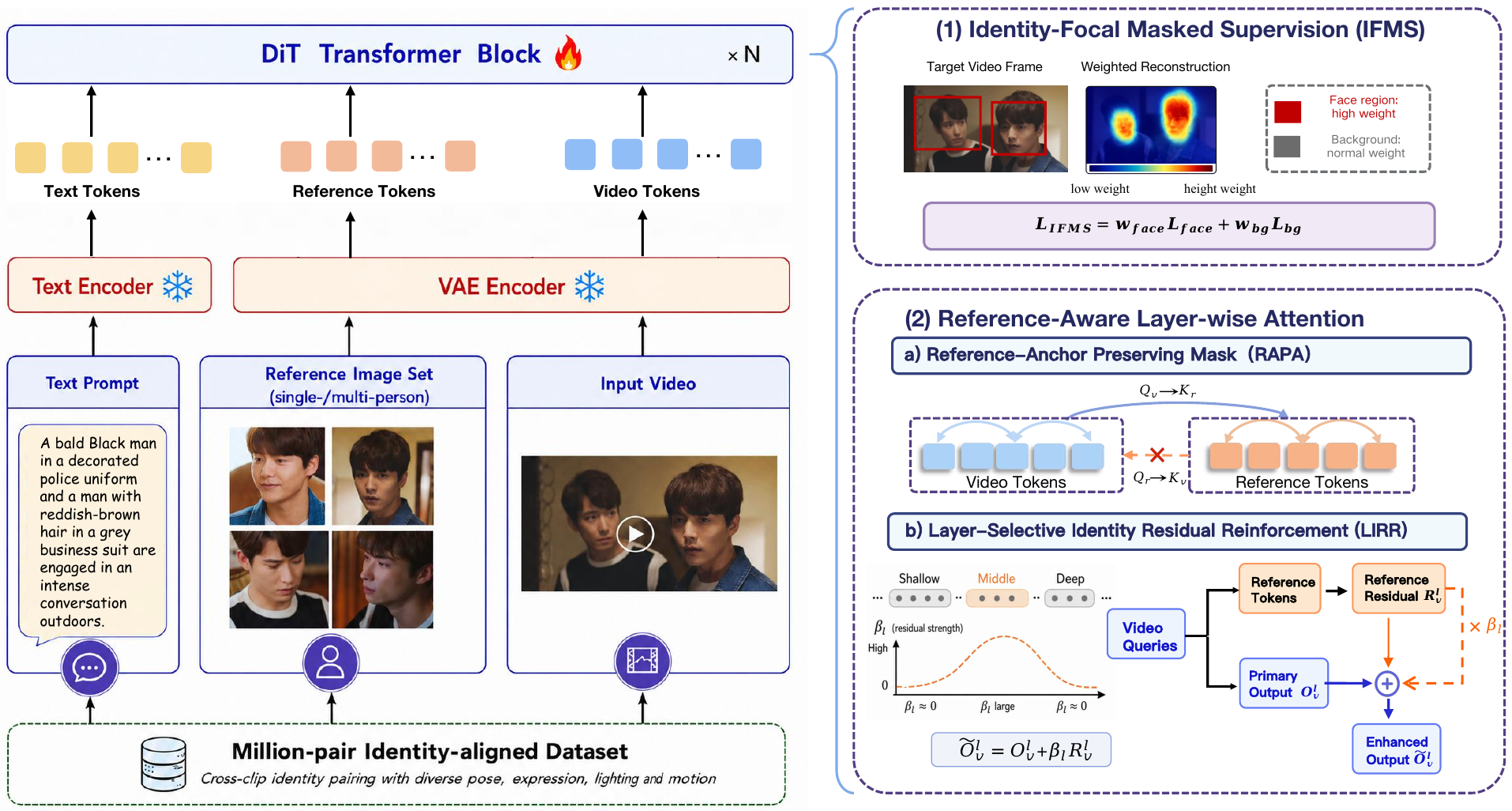}  
  \caption{ Overview of the Vera framework. Text prompts, reference image sets, and input videos are encoded into token sequences and processed by DiT Transformer blocks. Vera is built on a million-pair identity-aligned human image-video dataset and introduces two complementary designs: IFMS strengthens identity-aware spatial supervision, while RALA stabilizes reference-video interaction and enhances layer-aware identity readout. These designs enable identity-faithful and temporally coherent human S2V generation in both single- and multi-person scenarios.}
  \label{fig:pipeline}
\end{figure*}

In human-centric S2V generation, identity preservation is primarily determined by fine-grained facial details, such as face shape, local texture, and distinctive facial attributes. However, these identity-critical regions occupy only a small portion of each video frame. Standard reconstruction or flow-matching objectives treat all spatial locations uniformly, causing the training signal to be dominated by large non-face regions such as background, clothing, and body areas. As a result, the model may learn globally plausible appearance and motion while receiving insufficient supervision on the facial details that are essential for preserving a person's identity across frames.

To address this imbalance, we introduce \textbf{Identity-Focal Masked Supervision (IFMS)}, a region-aware training objective that reallocates the supervision strength toward identity-critical facial regions. For a given training video, we derive a face mask $M_{\mathrm{face}}$ from detected face bounding boxes, and define its complementary non-face region as the background mask $M_{\mathrm{bg}}$. We then construct a spatial weighting map:
\begin{equation}
    W =
    w_{\mathrm{face}} M_{\mathrm{face}}
    +
    w_{\mathrm{bg}} M_{\mathrm{bg}},
\end{equation}
where $w_{\mathrm{face}}$ and $w_{\mathrm{bg}}$ control the relative contribution of facial and non-face regions to the training objective. The background weight provides regular supervision for global appearance, motion, and scene context, while the face weight increases the optimization emphasis on identity-sensitive facial details. When $w_{\mathrm{face}}=w_{\mathrm{bg}}$, IFMS reduces to the vanilla uniform objective; using a larger relative face weight encourages the model to learn more discriminative identity cues without discarding contextual supervision.

We integrate this weighting map into the rectified-flow training objective. Given a noisy latent $z_t$, target velocity $u_t$, and model prediction $v_\theta(z_t,t,c)$ conditioned on $c$, the IFMS objective is defined as:
\begin{equation}
    \mathcal{L}_{\mathrm{IFMS}}
    =
    \frac{
    \sum
    W \odot
    \left(
    v_{\theta}(z_t,t,c)-u_t
    \right)^2
    }{
    \sum W
    },
\end{equation}
where $\odot$ denotes element-wise multiplication. Before loss computation, $W$ is downsampled to the latent resolution. The normalization by $\sum W$ keeps the overall loss scale stable across videos with different face sizes and numbers of visible subjects.




Crucially, for multi-subject videos, $M_{\mathrm{face}}$ is constructed as the union of all valid face regions, enabling IFMS to jointly preserve multiple identities. Compared to the vanilla uniform loss, IFMS provides amplified gradients for identity-critical areas while maintaining standard supervision for the global scene context. This simple yet effective modification introduces zero inference overhead, yet it compellingly encourages the model to synthesize temporally stable facial identities and significantly reduces subtitle-induced artifacts during human-centric video generation.

\subsection{Reference-Aware Layer-wise Attention}
\label{sec:3}
In the DiT backbone, reference tokens, video latent tokens, and motion-related condition tokens are jointly processed by self-attention. While this design allows video tokens to access identity cues from reference images, it also introduces two directional issues in reference-video interaction. First, reference queries may attend to video keys and values, causing the reference representation to be affected by dynamic denoising states, pose changes, background context, and occlusions. Second, the need for video tokens to read identity information from reference tokens is not uniform across all layers. This is consistent with recent analyses of DiT depth, which show that different DiT blocks play different roles in generation: shallow blocks mainly define global structures such as composition and outlines, deeper blocks refine details, and middle blocks serve an intermediate role~\cite{chen2024delta}. Our experimental analysis further shows that video-to-reference identity readout exhibits a clear layer-dependent pattern and is mainly concentrated in middle or specific DiT layers. Based on these observations, we introduce \textbf{Reference-Aware Layer-wise Attention (RALA)}, which consists of a reference-side protection mask and a video-side identity readout gate.
\paragraph{Reference-Anchor Preserving Mask.}
To keep reference tokens as clean identity anchors, we introduce a \textbf{Reference-Anchor Preserving Mask (RAPM)}. Let the visual tokens be ordered as $X=[X_v;X_r]$, where $X_v$ and $X_r$ denote video and reference tokens, respectively. RAPM is applied only to shallow layers:
\begin{equation}
M_{\mathrm{RAPM}}^{l}
=
\begin{cases}
\begin{bmatrix}
\mathbf{0} & \mathbf{0} \\
-\infty & \mathbf{0}
\end{bmatrix}, & l \leq L_s, \\[6pt]
\mathbf{0}, & l > L_s,
\end{cases}
\end{equation}
where rows and columns follow the order $[X_v;X_r]$, and $L_s$ denotes the number of shallow layers using RAPM. This mask only blocks reference-to-video attention in shallow layers, preventing reference tokens from being disturbed by noisy video states while preserving the video-to-reference pathway for identity readout. As shown in Fig.~\ref{fig:pipeline}, RAPM blocks reference-to-video attention in shallow layers, keeping reference tokens clean while preserving video-to-reference identity conditioning.

\paragraph{Layer-Selective Identity Residual Reinforcement.}
Beyond protecting reference tokens, we further introduce \textbf{Layer-Selective Identity Residual Reinforcement (LIRR)} to strengthen identity readout at layers where video tokens most actively absorb reference identity cues. For the $l$-th self-attention layer, we first compute the RAPM-masked attention map:
\begin{equation}
    A_l =
    \mathrm{softmax}
    \left(
    \frac{QK^\top}{\sqrt{d}}
    +
    M_{\mathrm{RAPM}}^{l}
    \right).
\end{equation}
From this attention map, we denote the video-side output as $O_v^l$ and its reference-derived identity component as $R_v^l$. Motivated by the observation that video tokens mainly acquire identity information in middle or specific DiT layers~\cite{chen2024delta}, we reinforce this identity residual as:
\begin{equation}
    \widetilde{O}_v^l =
    O_v^l + \beta_l R_v^l ,
\end{equation}
where $\beta_l$ controls the strength of identity residual reinforcement at layer $l$. We assign larger $\beta_l$ to the identified identity-sensitive middle layers and keep it small or zero for shallow and deep layers. This design enhances identity readout where it is most needed, while avoiding excessive reference dependence during early anchor protection or late detail refinement. Since LIRR operates on the post-attention output rather than the attention logits, it preserves the original attention competition among tokens and can be implemented without additional parameters.
\section{Experiments}
\label{sec:exp}
\begin{table}[t]
\centering
\caption{Quantitative comparison on the subject-to-video generation task. We report identity consistency, motion quality, naturalness, and prompt-following metrics. “↑” indicates higher is better.}
\label{tab:quantitative_comparison}
\setlength{\tabcolsep}{3pt}
\renewcommand{\arraystretch}{0.95}
\resizebox{\columnwidth}{!}{%
\begin{tabular}{lccccc}
\toprule
\multirow{2}{*}{Method}
& \multicolumn{2}{c}{Identity}
& \multicolumn{1}{c}{Motion}
& \multicolumn{1}{c}{Natural.}
& \multicolumn{1}{c}{Prompt} \\
\cmidrule(lr){2-3}
\cmidrule(lr){4-4}
\cmidrule(lr){5-5}
\cmidrule(lr){6-6}
& FaceSim-Arc$\uparrow$
& FaceSim-Cur$\uparrow$
& MotionSmooth.$\uparrow$
& NaturalScore$\uparrow$
& X-CLIP$\uparrow$ \\
\midrule
VACE-1.3B~\citet{jiang2025vace}    & 0.487 & 0.468 & \textbf{0.962} & 3.62 & 0.307 \\
VACE-14B~\citet{jiang2025vace}     & \underline{0.531} & \underline{0.507} & 0.945 & 3.65 & 0.311 \\
BindWeave~\citet{li2025bindweave}    & 0.512 & 0.488 & 0.939 & 3.66 & \textbf{0.343} \\
Phantom-1.3B~\citet{liu2025phantom} & 0.488 & 0.461 & 0.928 & \underline{3.71} & 0.333 \\
Phantom-14B~\citet{liu2025phantom}  & 0.495 & 0.472 & \underline{0.952} & 3.83 & 0.329 \\
\midrule
Ours         & \textbf{0.571} & \textbf{0.529} & 0.930 & \textbf{3.90} & \underline{0.340} \\
\bottomrule
\end{tabular}%
}
\end{table}

\subsection{Experiment Setup}

\noindent\textbf{Implementation Details.}
Vera is fine-tuned from Wan-2.2-TI2V-5B~\cite{wan2025wan}, a DiT-based foundation video generation model. We exclude the original T2V and I2V pre-training stages from this evaluation and focus on the human-centric S2V fine-tuning stage. All training videos are resampled to 16 fps and resized to $480 \times 832$. During training, each sample is conditioned on the text prompt and the corresponding human reference images. Face bounding boxes and subtitle masks are used to construct the bbox-aware masked loss, while the reference-preserving attention mask is applied in early transformer layers. We train the model with the Adam optimizer using a learning rate of $5 \times 10^{-6}$ and a global batch size of 32. All ablation studies are conducted under the same training settings unless otherwise stated.

\paragraph{Baselines.}
We compare Vera with strong subject-to-video baselines, including VACE~\cite{jiang2025vace}, Phantom~\cite{liu2025phantom}, and BindWeave~\cite{li2025bindweave}. For VACE~\cite{jiang2025vace} and Phantom~\cite{liu2025phantom}, we report both 1.3B and 14B variants to evaluate performance across model scales. VACE~\cite{jiang2025vace} provides a unified framework for reference-conditioned video creation, Phantom~\cite{liu2025phantom} learns subject consistency through cross-modal text-image-video alignment, and BindWeave~\cite{li2025bindweave} further leverages multimodal reasoning to model entities, attributes, and interactions. These methods represent competitive baselines for subject-consistent video generation and allow us to evaluate Vera on facial identity consistency, motion quality, naturalness, and prompt following.



\noindent\textbf{Benchmark and Evaluation Metrics.}
We evaluate Vera on a curated human-centric S2V benchmark containing 100 identities. Each identity is associated with two text prompts and corresponding reference images. For reliable evaluation, each text-image pair is generated with three random seeds, producing 600 videos in total.

We report metrics covering identity preservation, temporal motion, naturalness, and prompt following. For identity preservation, we compute FaceSim using ArcFace~\cite{deng2019arcface} and CurricularFace~\cite{huang2020curricularface}, denoted as FaceSim-Arc and FaceSim-Cur, respectively. For temporal motion, we report MotionSmoothness and MotionAmplitude following optical-flow-based evaluation protocols~\cite{bradski2000opencv}. For perceptual naturalness, we adopt NaturalScore. For prompt following, we use X-CLIP~\cite{ma2022x} to measure text-video semantic alignment. All metrics are reported with higher values indicating better performance.
\begin{figure*}[t]
  \centering

  \includegraphics[width=\linewidth]{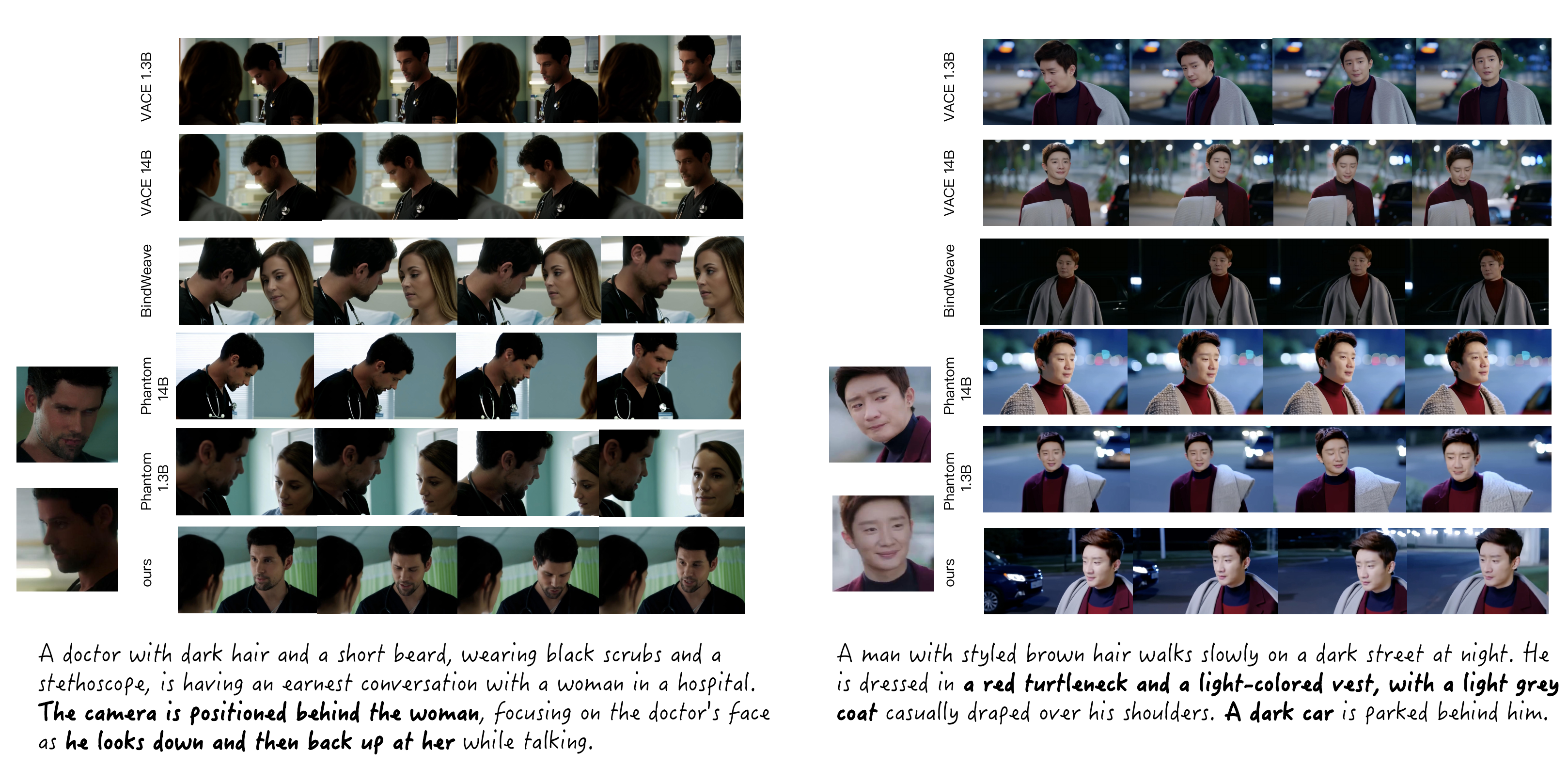}
  \caption{Qualitative comparison on single-person subject-to-video generation. Our method better balances identity consistency, appearance fidelity, and motion naturalness in conversational scenes.}
  \label{fig:single}

  \includegraphics[width=\linewidth]{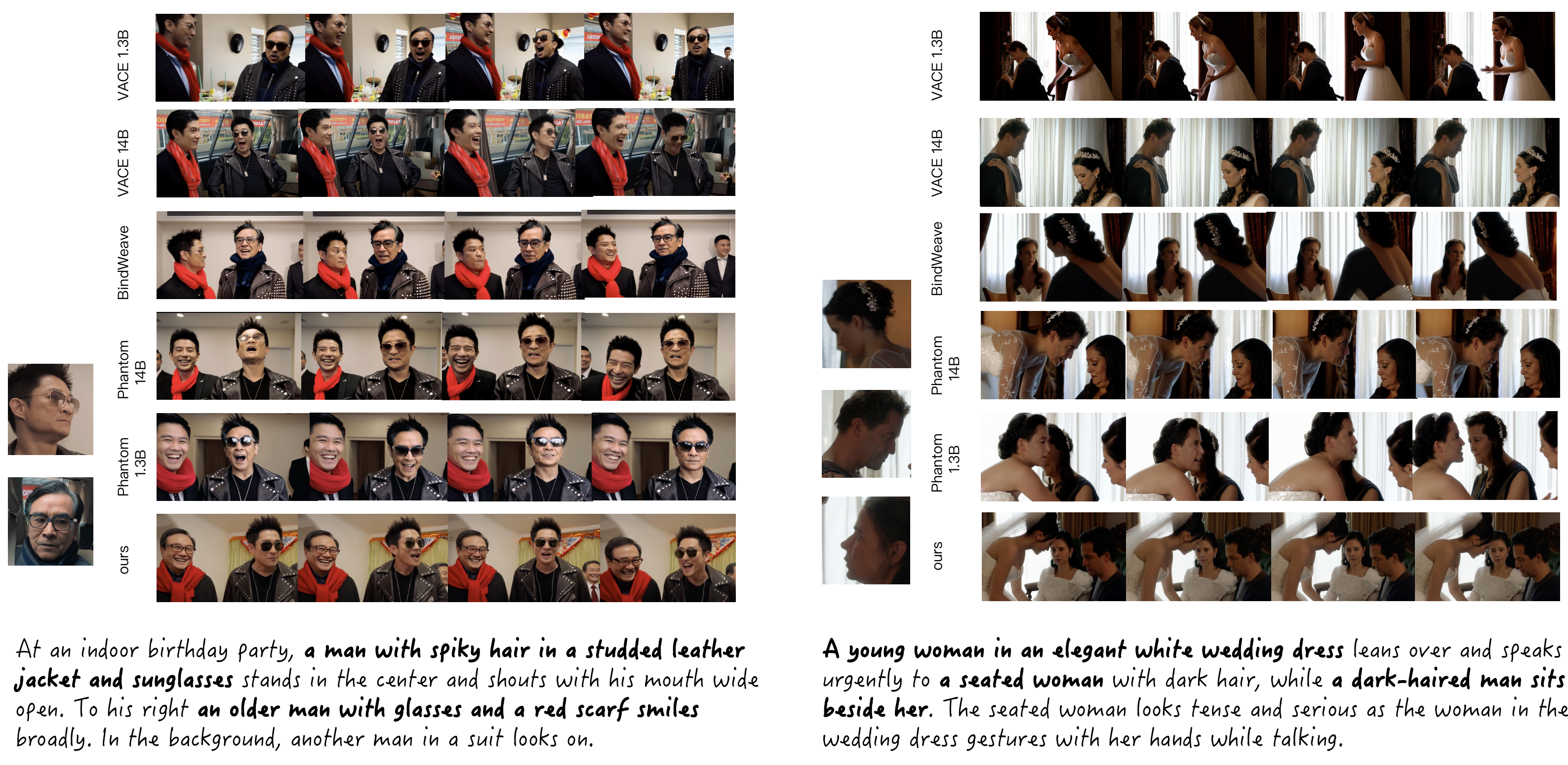}
  \caption{Qualitative comparison on multi-person subject-to-video generation. Our method better preserves individual identities, spatial layouts, and interaction consistency across complex multi-character scenes.}
  \label{fig:multi}

\end{figure*}

\subsection{Quantitative Comparisons}
We conduct quantitative comparisons with representative subject-to-video generation methods, including VACE~\cite{jiang2025vace}, BindWeave~\cite{li2025bindweave}, and Phantom~\cite{liu2025phantom}, as reported in Tab.~\ref{tab:quantitative_comparison}. We evaluate the generated videos from four complementary aspects: identity consistency, motion smoothness, perceptual naturalness, and prompt following.

Vera achieves the best identity consistency, with FaceSim-Arc and FaceSim-Cur scores of 0.571 and 0.529, respectively. Compared with the strongest baseline VACE-14B, Vera improves FaceSim-Arc from 0.531 to 0.571 and FaceSim-Cur from 0.507 to 0.529, indicating more reliable preservation of human identity across generated frames. In addition, Vera obtains the highest NaturalScore of 3.90 and a competitive X-CLIP score of 0.340, while maintaining reasonable temporal smoothness. Although VACE-1.3B achieves the highest MotionSmoothness score, its identity consistency remains substantially lower than that of Vera. These results suggest that Vera improves identity preservation and perceptual quality while maintaining competitive prompt alignment and temporal coherence, validating its effectiveness for identity-faithful human S2V generation.

\subsection{Qualitative Comparisons}

To demonstrate the effectiveness of Vera, we present representative single-person and multi-person subject-to-video comparisons in Fig.~\ref{fig:single} and Fig.~\ref{fig:multi}, including BindWeave~\cite{li2025bindweave}, Phantom-14B~\cite{liu2025phantom},  Phantom-1.3B~\cite{liu2025phantom}, VACE-1.3B~\citet{jiang2025vace} and VACE-14B~\citet{jiang2025vace} as baselines.
As shown in Fig.~\ref{fig:single}, baseline methods produce plausible videos but often exhibit identity drift or incomplete prompt alignment. In both the doctor conversation and night-street examples, they preserve only partial appearance or scene cues, while failing to consistently maintain the reference identity, fine-grained outfit details, and prompt-specified motion. In contrast, Vera better preserves identity and appearance while generating more natural, prompt-aligned motion.
Fig.~\ref{fig:multi} further demonstrates the advantage of Vera in multi-person scenarios. These cases require preserving multiple identities while maintaining correct role assignment, spatial layout, and interactions. Baseline methods often focus on only part of the scene, confuse character roles, or blend identities across subjects, especially in the birthday-party and wedding examples. By comparison, Vera more consistently preserves each referenced person, maintains clearer spatial relations, and generates interactions that better match the prompt.\par
Overall, Vera achieves balanced improvements across identity consistency, appearance fidelity, prompt alignment, and motion naturalness, with a particular advantage in identity preservation.

\subsection{Ablation study}
\begin{figure*}[t]
  \centering
  \includegraphics[width=\linewidth]{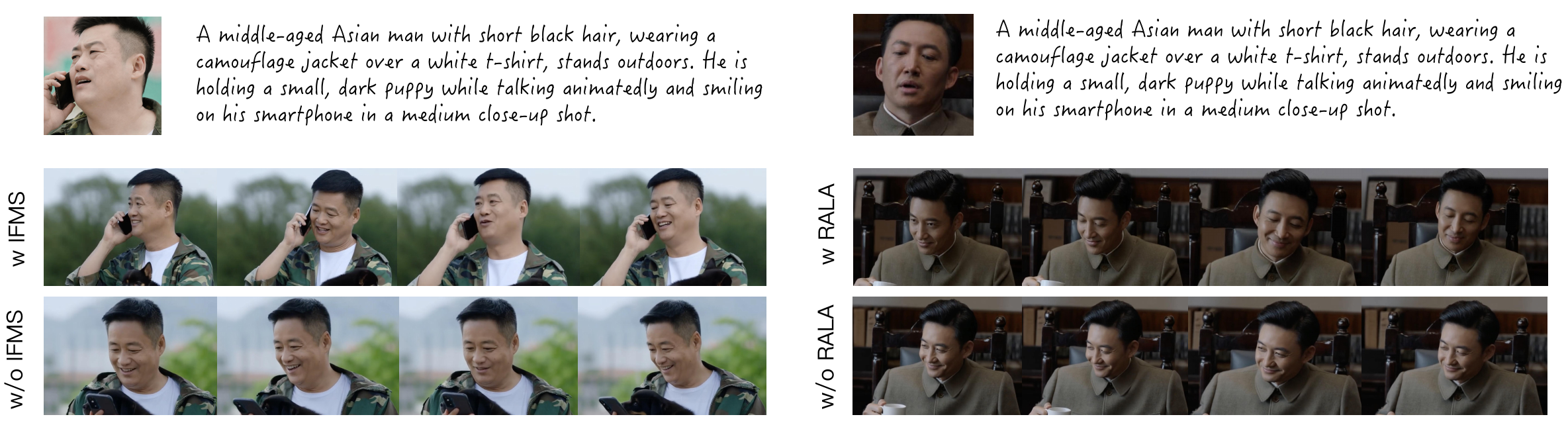}  
  \caption{ Qualitative ablation of IFMS and RALA. IFMS improves overall facial consistency, while RAPA further preserves fine-grained identity details by stabilizing reference anchors.}
  \label{fig:ablation}
\end{figure*}
\begin{table}[t]
\centering
\caption{Ablation study of IFMS and RALA. The full model achieves the best identity consistency, validating the contribution of each proposed component.
}
\label{tab:ablation}

\begin{tabular}{lccc}
\toprule
\multirow{2}{*}{Methods}
& \multicolumn{2}{c}{Identity Consistency}
& \multicolumn{1}{c}{Prompt Following} \\
\cmidrule(lr){2-3} \cmidrule(lr){4-4}
& FaceSim-Arc$\uparrow$
& FaceSim-Cur$\uparrow$
& X-CLIP$\uparrow$ \\
\midrule
w/o IFMS & 0.523 & 0.508 & 0.317 \\
w/o RALA & 0.545 & 0.513 & 0.309 \\
w/ All   & \textbf{0.571} & \textbf{0.529} & \textbf{0.340} \\
\bottomrule
\end{tabular}
\end{table}
Tab.~\ref{tab:ablation} reports the ablation results of our proposed components. Removing IFMS leads to clear degradation in identity consistency, with FaceSim-Arc dropping from 0.571 to 0.523 and FaceSim-Cur from 0.529 to 0.508. This verifies the importance of identity-aware spatial supervision for preserving discriminative human identity cues. Removing RALA also causes consistent drops across identity consistency and prompt following, suggesting that layer-aware reference-video interaction helps stabilize identity conditioning. With all components enabled, Vera achieves the best performance across all metrics, demonstrating that IFMS and RALA are complementary: IFMS strengthens identity-focused learning at the supervision level, while RALA improves reference conditioning and identity readout at the attention level.
As shown in Fig.~\ref{fig:ablation}, IFMS improves overall identity appearance by providing stronger supervision on identity-critical regions, leading to more consistent face shape, skin tone, and facial appearance. In comparison, RALA mainly benefits fine-grained identity preservation. Without RALA, the generated subject remains visually plausible but loses subtle reference-specific details such as the hairline, eyebrow shape, and skin texture. With RALA, these fine-grained identity details are better preserved across frames, suggesting that layer-aware reference-anchor preservation and identity residual reinforcement help maintain stable and detailed identity cues. Together, IFMS and RALA improve identity preservation from overall appearance consistency to fine-grained identity fidelity.

\section{Conclusion}
\label{sec:conclusion}
We introduce Vera, a unified human-centric S2V framework for identity-faithful single- and multi-person video generation. Unlike generic subject-consistent methods that mainly preserve coarse appearance, Vera focuses on maintaining stable person-level identity across poses, motions, and interactions. We construct a million-pair identity-aligned human image-video dataset through person-level cross-clip retrieval, providing diverse references and explicit identity correspondence. Built on this dataset, Vera introduces Identity-Focal Masked Supervision (IFMS) to strengthen identity-aware spatial supervision and Reference-Aware Layer-wise Attention (RALA) to stabilize reference-video interaction and enhance fine-grained identity readout. Experiments show that Vera improves human identity consistency, multi-person subject binding, and natural motion while reducing identity confusion and  excessive reference-image copying. These results highlight the importance of identity-aware supervision and reference-aware modeling for controllable human video generation.

\bibliographystyle{plainnat}
\bibliography{reference}

\appendix
\clearpage
\setcounter{page}{13}

\appendix
\renewcommand{\thesection}{\Alph{section}}
\renewcommand{\thesubsection}{\thesection.\arabic{subsection}}
\renewcommand{\theequation}{\arabic{equation}}

\begin{center}
    {\Large\bfseries Appendix}
\end{center}

\vspace{1em}

\noindent
\appentry{A\quad Preliminaries}{app:preliminaries}

\vspace{0.5em}
\appentry{B\quad Data Construction Pipeline}{app:data-construction}
\appsubentry{B.1\quad Video Segmentation and Face-Level Filtering}{app:video-segmentation}
\appsubentry{B.2\quad Intra-Clip Identity Clustering}{app:identity-clustering}
\appsubentry{B.3\quad Cross-Clip Retrieval and Diverse Reference Selection}{app:cross-clip-retrieval}

\vspace{0.5em}
\appentry{C\quad Experiment Settings}{app:experiment-settings}
\appsubentry{C.1\quad Implementation Details}{app:implementation-details}
\appsubentry{C.2\quad Evaluation Benchmark}{app:evaluation-benchmark}

\vspace{0.5em}
\appentry{D\quad Additional Ablation Results}{app:additional-ablation}
\appsubentry{D.1\quad Ablation Details on Identity-Focal Masked Supervision (IFMS)}{app:IFMS}
\appsubentry{D.2\quad Ablation on Reference-Aware Layer-wise Attention (RALA)}{app:RALA}

\vspace{0.5em}
\appentry{E\quad More Qualitative Results}{app:qualitative-results}

\vspace{0.5em}
\appentry{F\quad Dataset Statistics}{app:dataset-statistics}
\vspace{0.5em}
\appentry{G\quad Additional Statement}{app:additional-statement}
\appsubentry{G.1\quad Limitations and Future Work}{app:limitations}

\vspace{1.5em}

\section{Preliminaries}
\label{app:preliminaries}

In this section, we summarize the basic formulation of latent rectified-flow video generation and human-centric subject-to-video generation. We also introduce the attention notation used to describe the interaction between text, reference images, and video latent tokens in a DiT-based video generator.

\paragraph{Latent Rectified-Flow Video Generation.}
Given a clean video $X \in \mathbb{R}^{T \times C \times H \times W}$, a pretrained video VAE encoder $\mathcal{E}(\cdot)$ maps it into a compact latent representation:
\begin{equation}
    z_1 = \mathcal{E}(X),
\end{equation}
where $z_1 \in \mathbb{R}^{T' \times C' \times H' \times W'}$ denotes the clean video latent. We sample a Gaussian noise latent $z_0 \sim \mathcal{N}(0,I)$ with the same shape as $z_1$. Rectified flow defines a linear interpolation between the noise latent and the clean video latent:
\begin{equation}
    z_t = t z_1 + (1-t) z_0, \quad t \in [0,1].
\end{equation}
The corresponding target velocity is obtained by differentiating $z_t$ with respect to $t$:
\begin{equation}
    u_t = \frac{d z_t}{d t} = z_1 - z_0.
\end{equation}
A video generation model parameterized by $\theta$ is trained to predict this velocity field:
\begin{equation}
    \mathcal{L}_{\mathrm{RF}}
    =
    \mathbb{E}_{z_1, z_0, t, c}
    \left[
    \left\|
    v_{\theta}(z_t,t,c) - u_t
    \right\|_2^2
    \right],
\end{equation}
where $c$ denotes the conditioning signals, such as text prompts and reference images. During inference, the learned velocity field is integrated from noise to generate a video latent, which is then decoded back to pixel space by the VAE decoder.



\paragraph{Human-Centric Subject-to-Video Generation.}
Subject-to-video generation aims to synthesize videos conditioned on reference subjects and textual prompts. In this work, we focus on the human-centric setting, where the reference subjects are people and the goal is to preserve their identities across generated frames. Given a text prompt $p$ and a set of human reference images
\begin{equation}
    \mathcal{R} = \{R_i\}_{i=1}^{N},
\end{equation}
where $N$ denotes the number of reference identities, the model generates a video $\hat{X}$ that follows the prompt while preserving the identity of each referenced person.

For multi-person generation, the model must further maintain person-level correspondence between multiple references and multiple generated subjects. Specifically, each reference identity should be consistently associated with its corresponding textual role and generated subject throughout the video. This requires not only preserving fine-grained identity cues, but also avoiding identity-role misbinding, subject confusion, and attribute leakage in dynamic multi-person scenes.

\paragraph{Tokenized DiT Attention.}
In our DiT-based video generator, text prompts are encoded into text tokens and injected through cross-attention, while reference images and video latents are processed in the visual self-attention stream. We denote the reference tokens as $X_r$ and the video latent tokens as $X_v$. The concatenated visual tokens are written as:
\begin{equation}
    X = [X_r; X_v].
\end{equation}
For a self-attention layer, the tokens are projected into queries, keys, and values:
\begin{equation}
    Q = XW_Q,\quad K = XW_K,\quad V = XW_V.
\end{equation}
The self-attention output is computed as:
\begin{equation}
    \mathrm{Attn}(Q,K,V)
    =
    \mathrm{softmax}
    \left(
    \frac{QK^\top}{\sqrt{d}}
    \right)V.
\end{equation}

Under this formulation, video tokens can read identity information from reference tokens through the attention path from video queries to reference keys and values. For video queries, the attention output can be decomposed as:
\begin{equation}
    O_v = A_{v,v}V_v + A_{v,r}V_r,
\end{equation}
where $A_{v,v}$ denotes attention from video queries to video keys, and $A_{v,r}$ denotes attention from video queries to reference keys. The second term $A_{v,r}V_r$ represents the identity residual read from reference tokens.

However, standard self-attention also allows reference tokens to attend to video tokens, which may expose reference representations to noisy denoising states, pose changes, background context, and occlusions. This motivates our Reference-Aware Layer-wise Attention, which preserves reference tokens as stable identity anchors and selectively reinforces video-to-reference identity readout.

\section{Data Construction Pipeline}
\label{app:data-construction}

We build an identity-aligned image-video dataset to support human-centric subject-to-video generation. The key objective is to construct training pairs where each target video clip is paired with reference images of the same person, while avoiding trivial frame reconstruction and reference-specific shortcut copying. The pipeline consists of three stages: video segmentation and face-level filtering, intra-clip identity clustering, and cross-clip retrieval with diverse reference selection.

\subsection{Video Segmentation and Face-Level Filtering}
\label{app:video-segmentation}

Given a raw video, we first apply scene-cut detection to divide it into temporally coherent video clips:
\begin{equation}
    V \rightarrow \{V_i\}_{i=1}^{M},
\end{equation}
where each $V_i$ denotes a short clip with consistent scene content. For each clip, we uniformly sample frames to form a frame set $\mathcal{F}_i$. A pretrained face detector and recognition model is then used to detect faces and extract face embeddings from each sampled frame. Each detected face is represented as:
\begin{equation}
    d_{i,t}^{m} = (b_{i,t}^{m}, e_{i,t}^{m}, s_{i,t}^{m}),
\end{equation}
where $b_{i,t}^{m}$ is the face bounding box, $e_{i,t}^{m}$ is the face embedding, and $s_{i,t}^{m}$ is the detection confidence of the $m$-th face in frame $t$.

To remove unreliable samples, we filter clips and faces based on three criteria. First, faces with low detection confidence are discarded. Second, faces whose bounding-box area is too small relative to the frame are removed, since they provide weak identity cues. Third, clips that are too short are excluded to ensure sufficient temporal motion. This filtering stage produces a cleaner set of face detections for subsequent identity-level processing.

\subsection{Intra-Clip Identity Clustering}
\label{app:identity-clustering}

A single video clip may contain multiple people, transient background faces, or false detections. To obtain person-level identity units, we cluster all valid face embeddings within each clip according to cosine similarity. For clip $V_i$, the detected faces are grouped into identity clusters:
\begin{equation}
    \mathcal{C}_i = \{C_{i,k}\}_{k=1}^{N_i},
\end{equation}
where $C_{i,k}$ denotes the $k$-th identity cluster and $N_i$ is the number of valid identities in the clip. For each cluster, we compute an average identity embedding:
\begin{equation}
    \mu_{i,k} =
    \frac{1}{|C_{i,k}|}
    \sum_{e \in C_{i,k}} e .
\end{equation}
Clusters with too few occurrences are removed, as they usually correspond to transient faces, unreliable detections, or background people. The remaining clusters represent stable person identities appearing in the target clip. This step is especially important for multi-person videos, where different identities must be separated before constructing reference sets.

\subsection{Cross-Clip Retrieval and Diverse Reference Selection}
\label{app:cross-clip-retrieval}

Directly using frames from the target clip as references may encourage shortcut learning, where the model copies the background, pose, or frame layout rather than learning robust identity correspondence. To alleviate this issue, we construct reference images through cross-clip identity retrieval. For each identity cluster $C_{i,k}$ in target clip $V_i$, we search for identity-matched candidates from other clips of the same video source. Given another clip $V_j$ with identity cluster $C_{j,l}$, the identity similarity is computed as:
\begin{equation}
    \mathrm{sim}(C_{i,k}, C_{j,l})
    =
    \frac{\mu_{i,k}^{\top}\mu_{j,l}}
    {\|\mu_{i,k}\|_2 \|\mu_{j,l}\|_2}.
\end{equation}
We retain candidate clusters whose similarity falls within a moderate range:
\begin{equation}
    \tau_{\mathrm{low}}
    <
    \mathrm{sim}(C_{i,k}, C_{j,l})
    <
    \tau_{\mathrm{high}} .
\end{equation}
The lower threshold filters out mismatched identities, while the upper threshold removes near-duplicate clips that may lead to background leakage or shortcut copying of frame-specific visual cues.

From the retrieved candidates, we select a diverse set of reference images for each identity. Specifically, candidate faces are first ranked by identity similarity and detection quality. We then use visual features extracted by a pretrained image encoder to encourage diversity among selected references. Let $\phi(\cdot)$ denote the visual feature extractor and $\mathcal{S}_{i,k}$ denote the selected reference set for identity $C_{i,k}$. The selection favors references that are identity-consistent while being visually diverse:
\begin{equation}
    \mathcal{S}_{i,k}
    =
    \operatorname*{arg\,max}_{\mathcal{S} \subset \mathcal{A}_{i,k}}
    \left[
    \frac{1}{|\mathcal{S}|}
    \sum_{R \in \mathcal{S}}
    \mathrm{sim}(R, C_{i,k})
    +
    \lambda
    \frac{1}{|\mathcal{S}|^2}
    \sum_{R_a,R_b \in \mathcal{S}}
    \left\|
    \phi(R_a)-\phi(R_b)
    \right\|_2
    \right],
\end{equation}
where $\mathcal{A}_{i,k}$ is the candidate reference pool and $\lambda$ balances identity reliability and visual diversity. In practice, this selection encourages the reference set to cover variations in pose, expression, illumination, and background.

For multi-person clips, the above retrieval and selection process is performed independently for each identity cluster. As a result, each training sample is constructed as:
\begin{equation}
    \mathcal{D}_i =
    \left(
    V_i,\,
    p_i,\,
    \{\mathcal{S}_{i,k}\}_{k=1}^{N_i}
    \right),
\end{equation}
where $V_i$ is the target video clip, $p_i$ is the text prompt, and $\mathcal{S}_{i,k}$ is the reference image set for the $k$-th person in the clip. This produces identity-aligned training samples for both single-person and multi-person human S2V generation.

Finally, we perform visual verification by grouping the selected candidate references according to identity and marking different identities with distinct colors. This allows us to inspect whether all people appearing in the target video are correctly matched with their corresponding reference images. Through this pipeline, we obtain identity-consistent, diverse, and non-trivial image-video pairs, enabling scalable training for facial identity preservation and multi-person subject binding.

\section{Experiment Settings}
\label{app:experiment-settings}

\subsection{Implementation Details}
\label{app:implementation-details}
\textbf{Identity-focused loss and subtitle suppression.}
We apply a bbox-aware masked denoising loss in the VAE latent space to strengthen identity-related supervision. Face bounding boxes are first transformed according to the same resizing and cropping pipeline as the training videos, and then enlarged by 50\% in both width and height before being rasterized into latent-space masks. The loss weight is set to 2.0 inside the enlarged face regions and 1.0 elsewhere, corresponding to \texttt{face scale=1.0} and \texttt{face offset=1.0}. In addition, subtitle regions are excluded from the denoising loss using caption bounding-box masks, reducing the influence of identity-irrelevant textual artifacts.


\textbf{Layer-wise RALA schedule.}
For reference-aware attention control, we adopt a layer-wise identity readout schedule across the DiT backbone. The video-side attention output is decomposed into a video self-contribution and a reference-derived identity contribution. We keep the default reference contribution in early and late layers, and amplify the reference-derived identity term in the middle layers with $\gamma_l=1.5$. This is equivalent to adding an identity residual with coefficient $\beta_l=0.5$ in the middle layers.

\subsection{Evaluation Benchmark}
\label{app:evaluation-benchmark}

Existing subject-to-video benchmarks provide useful evaluation protocols for general subject consistency, but they are often insufficient for assessing human-centric identity preservation. In particular, human S2V requires the model to preserve fine-grained facial identity under diverse poses, expressions, motions, and viewpoints. In multi-person scenarios, the model must further bind each reference identity to the correct textual role and temporal trajectory. To provide a more targeted evaluation, we construct a human-centric S2V benchmark that covers both single-identity and multi-person generation.

Our benchmark consists of 100 carefully curated subject-text pairs, divided into two primary categories: single-identity and multi-person cases. The single-identity group contains 70 test cases, where each case includes one or multiple reference images corresponding to the same person. These cases are designed to evaluate identity preservation under diverse prompts, including portrait motion, walking, emotional expression, and pose variation. The multi-person group contains 30 test cases with varying complexity, including two-person, three-person, and more complex group scenarios. These cases are used to evaluate whether the model can preserve multiple identities simultaneously while maintaining correct role assignment and interaction consistency.

The benchmark is carefully designed to ensure broad coverage of human appearances, background settings, and interaction dynamics. We include subjects with diverse ages, genders, hairstyles, skin tones, clothing styles, and facial attributes. The prompts cover a wide range of environments, such as indoor rooms, hospitals, streets, parties, outdoor scenes, and low-light settings. We also include different interaction patterns, including face-to-face interaction, walking, sitting, gesturing, looking at another person, and multi-character social interactions. This diversity allows the benchmark to evaluate not only static identity similarity, but also identity stability under realistic motion and interaction conditions.

Each test case contains one or multiple reference images and is paired with a natural language prompt. The reference images are selected to provide clear identity cues while avoiding overly trivial near-duplicate views. The prompts are designed to ensure semantic clarity, visual plausibility, and sufficient motion description. This controlled design keeps the evaluation difficulty relatively consistent across cases, while still covering challenging scenarios such as pose variation, partial occlusion, complex backgrounds, and multi-person identity-role binding.

For reliable evaluation, each reference-prompt pair is generated with multiple random seeds, and the resulting videos are evaluated using metrics covering identity consistency, temporal motion, naturalness, and prompt following. This benchmark provides a focused testbed for measuring the ability of human-centric S2V models to generate identity-faithful, temporally coherent, and semantically aligned videos in both single-person and multi-person settings.

\section{Additional Ablation Results}
\label{app:additional-ablation}
\subsection{Ablation Details on Identity-Focal Masked Supervision (IFMS)}
\label{app:IFMS}

Tab.~\ref{tab:quantitative_comparison} studies the effect of different IFMS region weights. When the face weight is relatively small, e.g., $w_{\mathrm{face}}=1.5$, the model achieves reasonable motion smoothness but weaker identity consistency, suggesting that insufficient emphasis on facial regions limits the learning of identity-critical details. Increasing the face weight to $w_{\mathrm{face}}=2.0$ consistently improves identity preservation, achieving the best FaceSim-Arc and FaceSim-Cur scores of 0.571 and 0.529, respectively. It also obtains the best NaturalScore and X-CLIP score, indicating that stronger face-focused supervision not only improves identity fidelity but also benefits perceptual quality and prompt alignment.

However, further increasing the face weight to $w_{\mathrm{face}}=2.5$ leads to performance degradation across all metrics. This suggests that overly emphasizing facial regions may disturb the balance between local identity supervision and global video reconstruction, resulting in weaker motion smoothness and less natural generation. In addition, setting a lower background weight, i.e., $w_{\mathrm{bg}}=0.5$, slightly improves motion smoothness but reduces identity and prompt-following performance, showing that background and non-face regions still provide useful contextual supervision. Overall, $w_{\mathrm{face}}=2.0$ and $w_{\mathrm{bg}}=1.0$ provide the best trade-off between identity preservation, visual naturalness, prompt following, and temporal coherence.

\begin{table}[t]
\centering
\caption{Ablation study of IFMS region weights. We vary $w_{\mathrm{face}}$ and $w_{\mathrm{bg}}$ to evaluate the effect of face-focused supervision on identity consistency, motion smoothness, naturalness, and prompt following. The best value is bolded and the second-best is underlined.}
\label{tab:quantitative_comparison}
\setlength{\tabcolsep}{3pt}
\renewcommand{\arraystretch}{0.95}
\resizebox{\columnwidth}{!}{%
\begin{tabular}{ccccccc}
\toprule
\multirow{2}{*}{$w_{\mathrm{face}}$}
& \multirow{2}{*}{$w_{\mathrm{bg}}$}
& \multicolumn{2}{c}{Identity}
& \multicolumn{1}{c}{Motion}
& \multicolumn{1}{c}{Natural.}
& \multicolumn{1}{c}{Prompt} \\
\cmidrule(lr){3-4}
\cmidrule(lr){5-5}
\cmidrule(lr){6-6}
\cmidrule(lr){7-7}
& 
& FaceSim-Arc$\uparrow$
& FaceSim-Cur$\uparrow$
& MotionSmooth.$\uparrow$
& NaturalScore$\uparrow$
& X-CLIP$\uparrow$ \\
\midrule
1.5 & 0.5 & 0.546 & 0.508 & \textbf{0.936} & 3.79 & 0.331 \\
1.5 & 1.0 & 0.558 & 0.517 & \underline{0.933} & \underline{3.85} & \underline{0.337} \\
2.0 & 1.0 & \textbf{0.571} & \textbf{0.529} & 0.930 & \textbf{3.90} & \textbf{0.340} \\
2.5 & 1.0 & \underline{0.561} & \underline{0.520} & 0.924 & 3.82 & 0.332 \\
\bottomrule
\end{tabular}%
}
\end{table}

\subsection{Ablation on Reference-Aware Layer-wise Attention (RALA)}
\label{app:RALA}


\begin{table}[t]
\centering
\caption{Ablation study of the middle-layer identity residual weight $w_{\mathrm{mid}}$ in Layer-Selective Identity Residual Reinforcement (LIRR). We vary $w_{\mathrm{mid}}$ to evaluate how the strength of video-to-reference identity readout at identity-sensitive middle layers affects identity consistency, motion smoothness, naturalness, and prompt following. The best value is bolded and the second-best is underlined.}
\label{tab:lirr_wmid}
\setlength{\tabcolsep}{3pt}
\renewcommand{\arraystretch}{0.95}
\resizebox{\columnwidth}{!}{%
\begin{tabular}{cccccc}
\toprule
\multirow{2}{*}{$w_{\mathrm{mid}}$}
& \multicolumn{2}{c}{Identity}
& \multicolumn{1}{c}{Motion}
& \multicolumn{1}{c}{Natural.}
& \multicolumn{1}{c}{Prompt} \\
\cmidrule(lr){2-3}
\cmidrule(lr){4-4}
\cmidrule(lr){5-5}
\cmidrule(lr){6-6}
& FaceSim-Arc$\uparrow$
& FaceSim-Cur$\uparrow$
& MotionSmooth.$\uparrow$
& NaturalScore$\uparrow$
& X-CLIP$\uparrow$ \\
\midrule
1.0 & 0.544 & 0.509 & \textbf{0.935} & 3.81 & \textbf{0.343} \\
1.3 & 0.559 & 0.518 & \underline{0.933} & \underline{3.86} & \underline{0.341} \\
1.5 & \textbf{0.571} & \textbf{0.529} & 0.930 & \textbf{3.90} & 0.340 \\
1.8 & \underline{0.562} & \underline{0.521} & 0.926 & 3.83 & 0.335 \\
2.0 & 0.550 & 0.513 & 0.921 & 3.77 & 0.329 \\
\bottomrule
\end{tabular}%
}
\end{table}

Tab.~\ref{tab:lirr_wmid} analyzes the effect of the middle-layer identity residual weight $w_{\mathrm{mid}}$ in LIRR. When $w_{\mathrm{mid}}$ increases from 1.0 to 1.5, identity consistency improves steadily, with FaceSim-Arc increasing from 0.544 to 0.571 and FaceSim-Cur from 0.509 to 0.529. This indicates that moderately strengthening video-to-reference identity readout at identity-sensitive middle layers helps video tokens absorb more reliable reference identity cues. Meanwhile, the NaturalScore also improves from 3.81 to 3.90, suggesting that enhanced identity readout can improve perceptual quality without introducing obvious artifacts.

However, further increasing $w_{\mathrm{mid}}$ to 1.8 or 2.0 leads to consistent degradation across identity, motion, naturalness, and prompt-following metrics. This suggests that overly strong reference residual reinforcement may make the generation depend too heavily on the reference tokens, reducing the model's flexibility in motion synthesis and semantic alignment. In particular, MotionSmoothness gradually decreases as $w_{\mathrm{mid}}$ becomes larger, indicating a trade-off between identity reinforcement and temporal dynamics. Overall, $w_{\mathrm{mid}}=1.5$ achieves the best identity consistency and naturalness while maintaining competitive motion smoothness and prompt following, providing the best balance for layer-selective identity residual reinforcement.

\section{More Qualitative Results}
\label{app:qualitative-results}
\begin{figure*}[t]
  \centering
  \includegraphics[width=\linewidth]{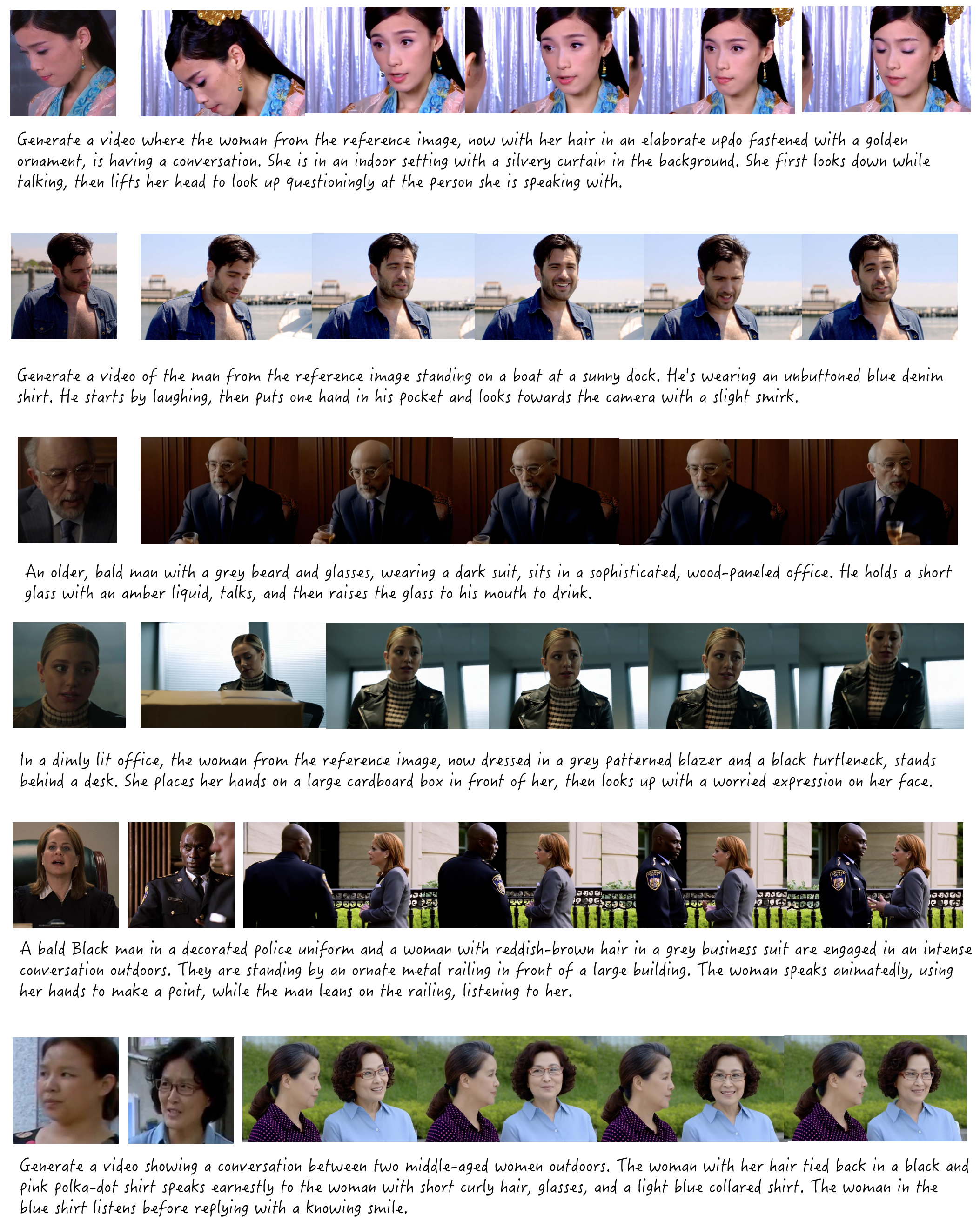}  
  
  \caption{ More qualitative results of Vera. Each case contains the input reference image(s), generated video frames, and the corresponding text prompt. The first four rows show single-person generation under diverse poses, expressions, motions, and scene contexts. The last two rows show multi-person generation, where Vera preserves multiple identities while maintaining correct role assignment and interaction consistency. These examples demonstrate the ability of Vera to generate identity-faithful, temporally coherent, and prompt-aligned human videos. }
  \label{fig:more_qualitative}
\end{figure*}
We provide additional qualitative results of Vera in Fig.~\ref{fig:more_qualitative}. These examples further demonstrate the effectiveness of our method in human-centric subject-to-video generation under diverse identities, appearances, motions, and interaction scenarios. Given one or multiple human reference images and a natural language prompt, Vera consistently generates temporally coherent videos while preserving the identity of the referenced subjects.
The first four rows show single-person generation results with diverse appearances and motion patterns, including head movement, facial expression changes, speaking, laughing, and object interaction. Vera maintains fine-grained identity cues such as face shape, hairstyle, skin tone, and facial attributes, while allowing the generated subjects to perform prompt-specified actions under different backgrounds and lighting conditions. The last two rows present multi-person generation cases, where the model is required to preserve multiple identities simultaneously and associate each person with the correct textual role. These results show that Vera can maintain person-level identity consistency and reduce identity confusion during dynamic interactions such as outdoor conversations.
Together, these results indicate that Vera generalizes well across varied human appearances, scene contexts, and motion patterns. The model not only preserves identity-critical facial details across frames, but also follows textual instructions and maintains coherent interactions in both single-person and multi-person settings.
\section{Dataset Statistics}
\label{app:dataset-statistics}
Following the data construction pipeline described in
Sec.~\ref{sec:1}, we obtain 1,001,891 identity-aligned
human image--video pairs covering both single-person and multi-person
scenarios. Each pair is constructed through video segmentation,
face-level identity clustering, and cross-clip reference retrieval.
This design preserves person-level identity correspondence while
introducing natural variations in pose, expression, illumination,
background, and motion.

Fig.~\ref{app:qualitative-results} summarizes the composition of the
resulting dataset. For gender-related annotations, 46.2\% of the
samples contain man/male descriptions, 27.2\% contain woman/female
descriptions, and 26.5\% are categorized as mixed or ambiguous.
Among samples with explicit age descriptions, middle-aged and young
adult subjects account for 43.6\% and 21.5\%, respectively, while
child/teen and elderly subjects account for 3.3\% and 1.5\%.
The dataset also contains diverse human-centric semantics, including
looking (84.0\%), speaking (79.3\%), walking (78.4\%), serious
expressions (66.1\%), standing (58.5\%), and sitting (35.4\%).
These semantic labels are non-exclusive and may co-occur within the
same sample. Overall, the dataset provides broad coverage of human
appearance, pose, expression, motion, and interaction patterns,
supporting identity-faithful generation under diverse conditions.
\label{app:qualitative-results}
\begin{figure*}[t]
    \centering

    \begin{subfigure}[t]{0.56\linewidth}
        \centering
        \includegraphics[width=\linewidth]{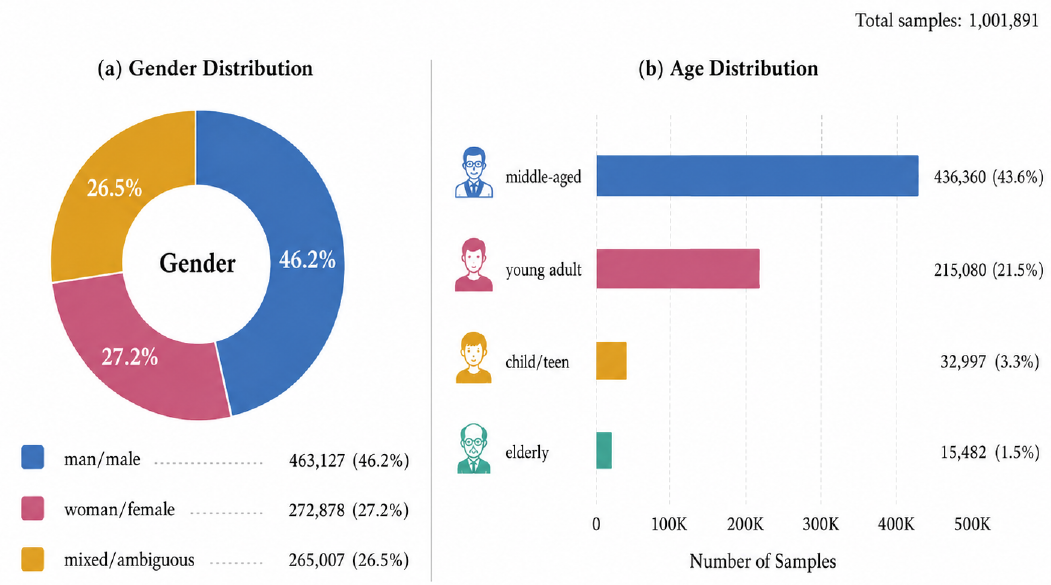}
        \caption{Distribution of demographic attributes in the identity-aligned dataset.}
        \label{fig:person_crop}
    \end{subfigure}
    \hfill
    \begin{subfigure}[t]{0.30\linewidth}
        \centering
        \includegraphics[width=\linewidth]{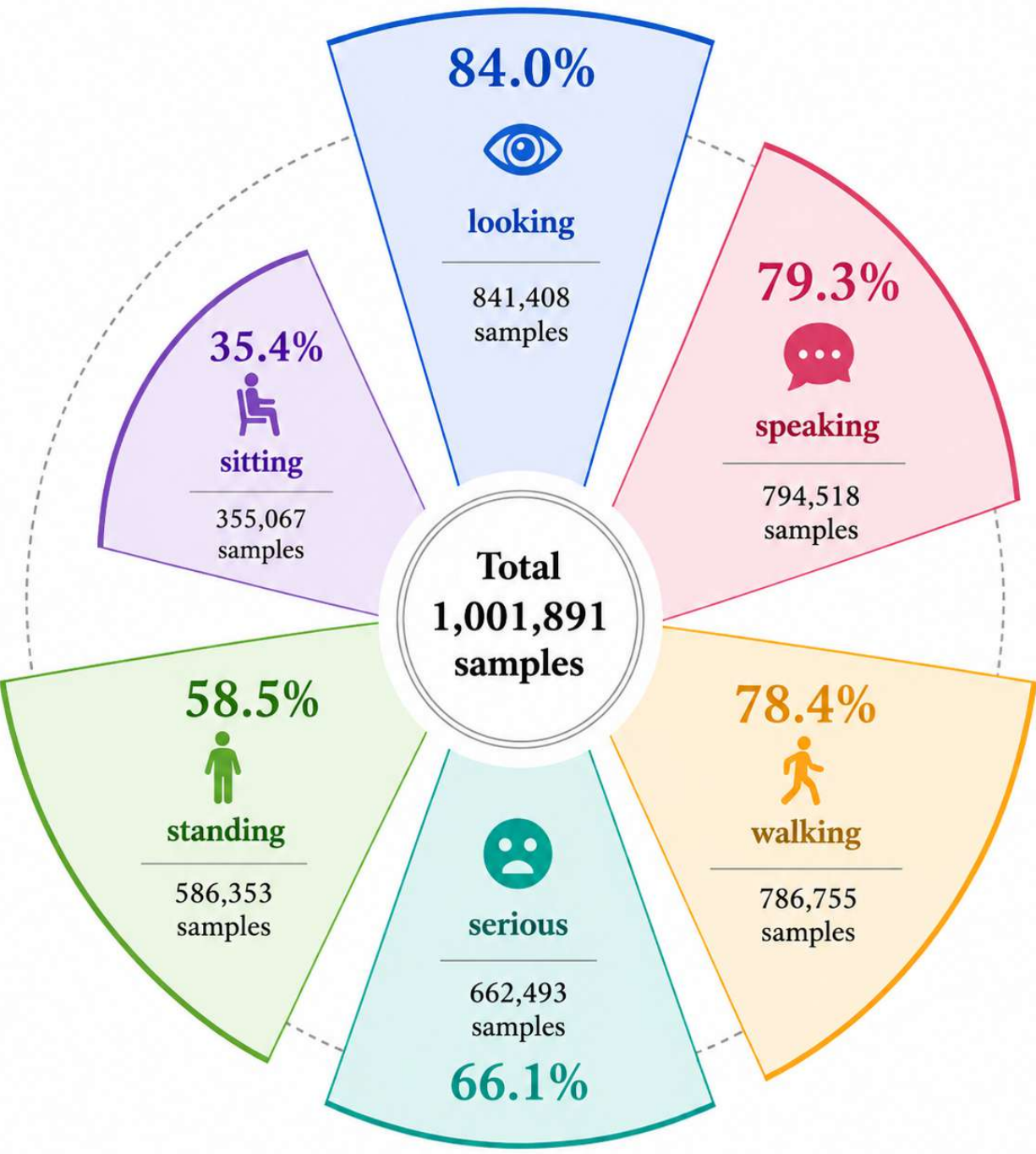}
        \caption{Frequencies of representative human-centric actions,
        poses, and expressions.}
        \label{fig:face_crop}
    \end{subfigure}

    \caption{
    Statistics of the identity-aligned human image--video dataset.
    The dataset contains 1,001,891 samples constructed through
    face-level identity clustering and cross-clip reference retrieval,
    covering both single-person and multi-person scenarios.
    }
    \label{fig:more_qualitative_results}
\end{figure*}

\section{Additional Statement}
\label{app:additional-statement}

\subsection{Limitations and Future Work}
\label{app:limitations}
Although Vera improves identity preservation for human-centric subject-to-video generation, several limitations remain. First, the current framework mainly supports image references and text prompts as generation conditions. This limits its applicability in scenarios that require richer control signals, such as video references, audio cues, motion trajectories, or 3D spatial information. Incorporating these additional modalities could provide more explicit guidance for facial expressions, body motion, speech-related dynamics, and spatial interactions, enabling more controllable and realistic human video generation.

Second, while Vera improves identity-role binding in multi-person scenarios, scaling to scenes with more people remains challenging. As the number of reference identities increases, the model must maintain more complex correspondences among reference images, textual roles, spatial positions, and temporal trajectories. Crowded scenes or visually similar identities may still lead to subject confusion or attribute leakage.

Third, our current framework is mainly evaluated on short video clips. Extending identity-faithful generation to longer videos remains an open problem, since small identity errors may accumulate over time and the model must preserve consistent identities, roles, and interactions across longer temporal contexts.

Future work will explore multi-modal conditioning with video, audio, motion, and 3D cues, stronger person-level binding mechanisms for crowded scenes, and long-video generation strategies that maintain identity consistency over extended durations.


\end{document}